\documentclass[journal]{IEEEtran}

\ifCLASSINFOpdf
\else
\fi
\usepackage{hyperref}

\usepackage{graphicx}
\usepackage{amsmath,amssymb,amsfonts}
\usepackage{algorithmic}
\usepackage{graphicx}
\usepackage{textcomp}
\usepackage{multirow}
\usepackage{makecell}
\usepackage{subfig}
\usepackage{colortbl}
\usepackage{enumitem}
\usepackage[dvipsnames]{xcolor}
\definecolor{mygray}{gray}{0.8}
\newcolumntype{a}{>{\columncolor{mygray}}c}
\begin{document}
%


\title{A Multi-stage deep architecture for summary generation of soccer videos}
%
%
%

\author{Melissa~Sanabria,
        Frédéric~Precioso,
        Pierre-Alexandre~Mattei,~and~Thomas~Menguy
}

\maketitle

\begin{abstract}

Video content is present in an ever-increasing number of fields, both scientific and commercial. Sports, particularly soccer, is one of the industries that has invested the most in the field of video analytics, due to the massive popularity of the game and the emergence of new markets (such as sport betting markets). Previous state-of-the-art methods on soccer matches video summarization rely on handcrafted heuristics to generate summaries which is poorly generalizable, but these works have yet proven that multiple modalities help detect the best actions of the game. On the other hand, machine learning models with higher generalization potential have enter the field of summarization of general-purpose videos, offering several deep learning approaches. However, most of them exploit content specificities that are not appropriate for sport whole-match videos. Although video content has been for many years the main source for automatizing knowledge extraction in soccer, the data that records all the events happening on the field has become lately very important in sports analytics, since this event data provides richer context information and requires less processing. Considering that in automatic sports summarization, the goal is not only to show the most important actions of the game, but also to reproduce the storytelling of the whole match with as much emotion as the one evoked by human editors, we propose a method to generate the summary of a soccer match video exploiting both the audio and the event metadata of the entire match. The results show that our method can detect the actions of the match, identify which of these actions should belong to the summary and then propose multiple candidate summaries which are similar enough but with relevant variability to provide different options to the final editor. Furthermore, we show the generalization capability of our work since it can transfer knowledge between datasets from different broadcasting companies, from different competitions, acquired in different conditions, and corresponding to summaries of different lengths.
\end{abstract}


%
\IEEEpeerreviewmaketitle

\section{Introduction}
\label{sec:intro}


Sports, particularly soccer, is one of the sectors that has invested the most in the video analysis field, due to the massive popularity of the game and the legalization of sport bet market. In a professional league such as Premier League, for example, with 10 matches per weekend, video from every stadium and multiple camera angles can quickly add up to dozens of hours of footage. And some companies manage the broadcasting of several competitions at the same time. In addition, the fans expect the summaries and highlights to be available as soon as the match is finished. Yet, most of the process for producing summary videos in broadcasting companies is still labor-intensive, time-consuming and not scalable. 

For many years automatic video summarization in sports mainly has relied on hand-crafted heuristics using video content as the main source. However, event data acquired in live during the matches is a very important source of information which very recently started to be exploited in the machine learning field for sports. The event data provides specific details of all the events happening on the field like the type of action, the position on the field, the players involved, the part of the body with which the player touched the ball, etc. Processing event data is significantly faster than video frames since a soccer match contains in average around 1700 events compared with more than 130k frames. In addition, video frames lead to several issues like occlusions, resolution, or subjectivity (when considering broadcast videos since they have already been edited in live by the producer), etc. 

Even though event data provides a lot of advantages over video, we cannot neglect the importance of the different modalities in sports. We exploit the audio features since they help to identify the excitement of the commentators and the crowd which is very important to detect the most relevant actions for a summary. 

One of the main challenges in sports summarization is the subjectivity since there is not a unique and perfect ground-truth summary for a match, it might depend on the platform where the video will be published, the league, the country, the length constraint, etc. We propose a method to provide several candidate summaries leaving the subjectivity decisions to the editors. 

\textbf{Contributions:}

\noindent$\bullet$ We propose to combine event data with audio features to automatically generate video summaries of soccer matches.

\noindent$\bullet$ Based on the variability among similar actions and, on the similarity between the actions that belong to the summary and the ones that do not, we propose an LSTM-Multiple Instance Learning schema for sequential data to generate action proposals.

\noindent$\bullet$ We describe a Hierarchical Multimodal Attention (HMA) mechanism that, unlike the state-of-the-art methods, in the first stage learns the importance of each modality (event data and audio) at the event level and in the second stage learns the importance of each event inside the action. HMA outperforms by 7\% the state-of-the-art methods in multimodal attention and by 15\% the ``soccer baselines''.

\noindent$\bullet$ We present a first-of-kind method for automatically generating multiple summaries of a soccer match by sampling from a ranking distribution. We provide different options to the editor solving two relevant issues in sports summarization, subjectivity and time constraints. In terms of F-score, our method outperforms by 6\% the state-of-the-art. And in terms of generalization, it outperforms by 9\% the state-of-the-art in prediction without fine-tuning the model to adapt it to a different competition. 

\begin{figure*}[!ht]
	\includegraphics[width=\textwidth]{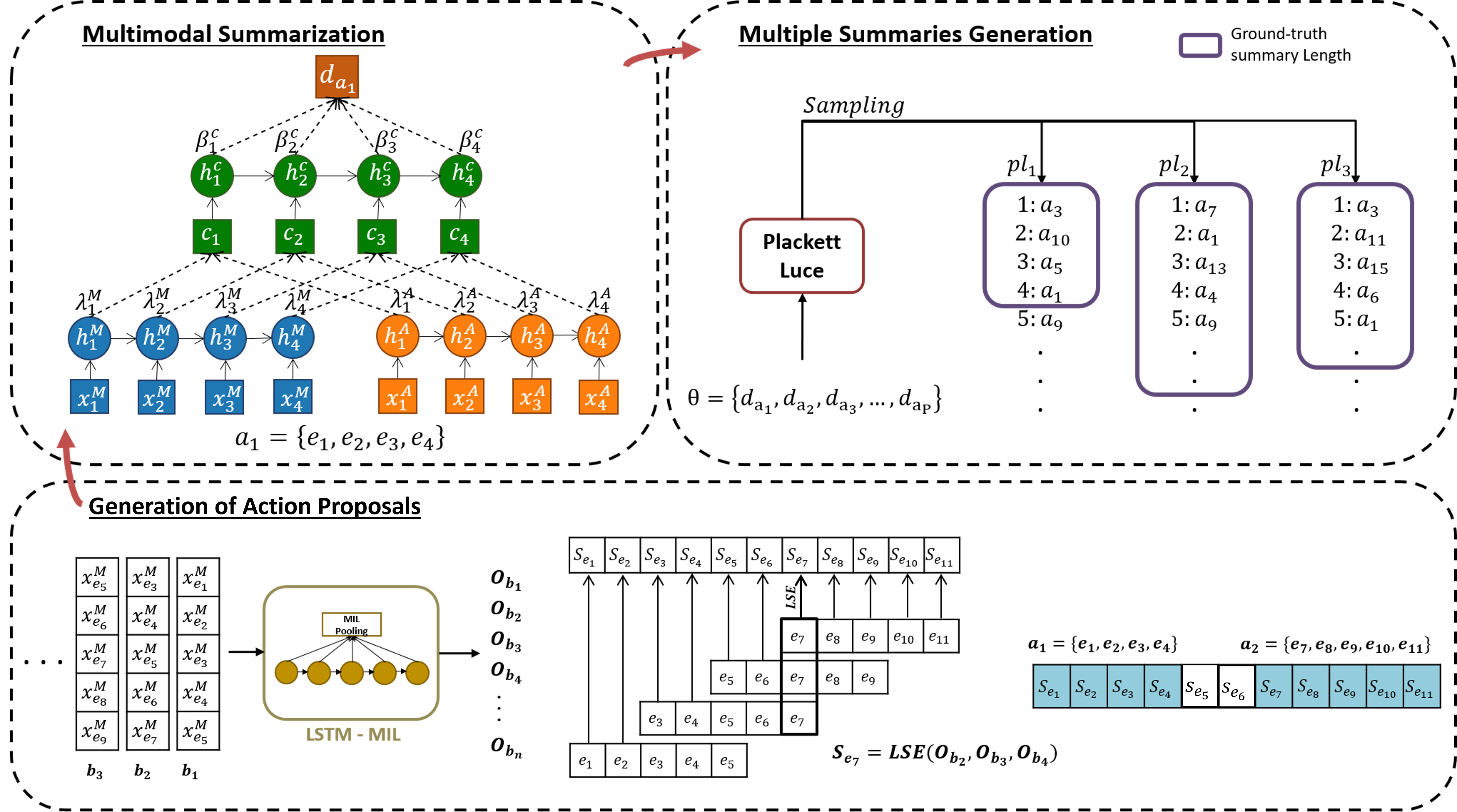}
	\caption{Our proposed approach for soccer video summarization. The \textit{Generation of Action Proposals} stage takes as input events $e_f$ grouped into bags $b_n$ and outputs action proposals. Each bag $b_n$ is processed by an LSTM Multiple Instance Learning (MIL) network, then Log-Sum-Exp (LSE) function is applied on all the predicted values of each event to finally group the consecutive positive events into action proposals. The \textit{Multimodal Summarization} stage uses a Hierarchical Multimodal Attention (HMA) that takes as input the action proposals $a_p$ and outputs the likelihood of each of the actions to be part of the summary. The \textit{Multiple Summaries Generation} stage takes as input the importance of each action $\theta_p$ as parameters for a Plackett- Luce distribution to generate multiple summaries of the same length as the ground-truth summary.}
	\label{fig:approach_diagram}
\end{figure*}

First, we will explore the state of the art of the different areas where our work has an impact. Then, in section \ref{sec:challenge} we will explain the differences between general-purpose videos and sports videos, and describe how the event data might be a better source of information than video content. After, section \ref{sec:our_approach} describes the three stages of our model shown in Figure \ref{fig:approach_diagram}: generation of actions proposals, multimodal summarization and multiple summaries generation. Section \ref{sec:experiments} contains the details of the datasets, architectures and features used to perform the experiments. And finally, section \ref{sec:results} shows the results of each stage of our method and their respective comparisons.

\section{Related Work}
\label{sec:related_work}
As the main goal of our work is to generate video summaries of soccer matches using multimodal information such as event data and audio, we will describe the state of the art of four different areas where our work has an impact: summarization of general-purpose videos, video summarization of sports, video summarization using different modalities and we describe some methods that also exploit event data.

Works on \textbf{Summarization of general-purpose videos} are not usually dedicated to sports domain and all the specific challenges that this task entails. Some methods \cite{rochan2018video, huang2019novel, yuan2019spatiotemporal, ji2020deep} need to load the whole video in memory as input sample, which is not feasible for a 90-minute video such as a soccer match. Several approaches are based on the maximization of diversity, trying to minimize the number of similar shots \cite{wang2019stacked,zhang2016video,rochan2019video, elfeki2019video, zhou2018deep, zhao2019property}, or assume that videos of similar topics share similar structures \cite{zhang2016summary,panda2017collaborative}, or even use a mixture of objectives like uniformity, interestingness and representativeness to identify the most appealing moments \cite{gygli2015video,li2017general}. However, the maximization of diversity is not an optimal approach for sports summarization. For instance, in soccer actions like goals, corners or free-kicks are visually very similar since they are located in the same area of the field. And this situation holds for many sports. On the other hand, some approaches have shown good results on general-purpose datasets using recurrent neural networks without any assumption on the content. Zhang et al. \cite{zhang2016video} propose a bidirectional LSTM followed by a Multi-Layer Perceptron and Zhao et al. \cite{zhao2017hierarchical, zhao2018hsa} describe a hierarchical LSTM to help the model to handle long-range structural dependencies. However they do not take into account the multiple modalities available for videos or the big difference between the summary length and the original video (see Section \ref{sec:challenge}). Recent successes of Generative Adversarial Networks have led to several works based on unsupervised approaches for video summarization \cite{rochan2019video,he2019unsupervised, apostolidis2020unsupervised}. For a more extensive list of the methods on video summarization using deep neural networks, the reader can read the survey provided by Apostolidis et al. \cite{apostolidis2021video}.

Early works in \textbf{Video Summarization for Sports} mainly rely on hand-crafted heuristics. They exploit context characteristics like lines, goal mouth, cinematographic properties like the camera motions, slow motion or zooming, and also specific edition patterns like replays, to select representative parts of the video \cite{ekin2003automatic,eldib2009soccer,tavassolipour2014event}. Recently, machine learning has taken an important role in soccer action detection. Liu et al. \cite{liu2017soccer} use 3D convolutional networks to classify the different clips of soccer videos, and Agyeman et al. \cite{agyeman2019soccer} use features extracted from 3D convolutions to then train an LSTM for action classification. Javed et al. \cite{javed2019replay} mix heuristics knowledge based on the replay information and an extreme learning machine to detect key-events. The main limitations in the state-of-the-art of video sport summarization is the lack of standardization in the evaluation process and in the assumptions based on heuristics. Most of the methods do not provide evaluation with commonly used summarization metrics, focusing instead on the accuracy of detecting the most important actions like goals.

Several works in the state-of-the-art tackle the problem as a \textbf{Multimodal Sports Summarization} task instead of considering only the video, since multiple modalities play an important role to choose the best moments of sports videos. Some methods propose to use the interactions on social networks like the tweet streams during the game \cite{tang2012epicplay,huang2018event}. Tang et al. \cite{tang2018autohighlight} use deep learning to classify soccer actions from the text timeline found in several web pages. Other methods exploit audio features \cite{rui2000automatically,baijal2015sports} since they help to identify the excitement of the commentators and the crowd, and sometimes the ball hit like for tennis or baseball. Several methods merge different modalities like the sound energy, the score, camera motions, players' reactions, referee whistle, etc \cite{bettadapura2016leveraging,merler2017automatic,merler2018automatic,shukla2018automatic}. In a previous work \cite{sanabria2021hierarchical}, we have proposed fully automatic soccer match summarization using event metadata and audio features. Our current work exploits the knowledge from building that first model in order to design a system answering more difficult questions: (i) instead of targeting a single (possibly imperfect) output, it provides multiple summaries per match to leave space for human editor choices; (ii) it allows to create length-constrained summaries; (iii) it can learn from one competition and transfer the knowledge to a new dataset that is very different to the one it was originally trained with in terms of content (visual, audio, etc) and, of targeted summary length.

\textbf{Event data} is produced by human annotators in live during the match, and is another modality from which we can extract valuable information. While our method uses this type of data for summarization, several approaches use it for other tasks like analyzing advantage of playing on the home field \cite{lucey2013assessing}, recognizing teams \cite{bialkowski2014identifying}, automatically discovering patterns in offensive strategies \cite{van2015automatically,gyarmati2015automatic}, predicting passes \cite{vercruyssen2016qualitative}, detecting tactics \cite{decroos2018automatic}, predicting the chance to score the next goal \cite{liu2018deep}, evaluating the performance or contributions of the players \cite{pappalardo2019playerank,decroos2019actions,bransen2018measuring} and modeling ball possession \cite{chacoma2020modeling}. These metadata can now more and more be found either on websites directly managed by the companies producing them (Prozone, GeniusSports,  Opta, WyScout, and others) or through open data sources \cite{pappalardo2019public, bergmann2013linked}. In all existing works, event data has been used to provide sports analytics, to the best of our knowledge our work is the first to consider event data as a core modality to be combined with raw signal data such as audio or video.


\section{The challenge of summarizing sport videos}
\label{sec:challenge}

\subsection{General-Purpose Videos vs Sport Videos}

As aforementioned the field of video summarization of general-purpose content has progressed rapidly providing promising results. However, there are several differences between summarizing general-purpose videos and sport videos. In terms of benchmark datasets \cite{gygli2014creating, song2015tvsum, de2011vsumm, ovp2011, zeng2016generation}, the length of general-purpose videos varies from 6 to 10 minutes and the summary length is around $15\%$ of the original video (56 to 90 seconds). On the other hand, for sport videos a match can vary from one to several hours and in sports like soccer the summaries are about few minutes, for instance a medium-large summary can be 5 minutes long which represents less than 6\% of the original video of a soccer game. 

\begin{table}[!ht]
\small
\begin{center}
\caption{Comparison between General Purpose dataset and Sports videos. Durations are in seconds. \textit{Soccer matches} represent a sample of 100 soccer matches from Premier League competition, the action duration is taken from the actions posted by official broadcasters. Action ratio is the ratio between the action duration and the video duration.}
\label{tab:datasets_comparison}
\begin{tabular}{|l|c|c|c|c|}
\hline
\thead{Dataset} & \thead{Number \\of videos} & \thead{Action \\duration} & \thead{Video \\duration} &\thead{Action \\Ratio}\\
\hline
ActivityNet \cite{caba2015activitynet} & 20K & 49.21 & 116.7 & 49.12\\
\hline
THUMOS-14 \cite{jiang2014thumos} & 412 & 4.6 & 213 & 31.01\\
\hline
HACS \cite{zhao2019hacs} & 1.5M & 40.6 & 156 & 30.84\\
\hline
Soccer matches & 100 & 30.84 & 6300 & 22.16\\
\hline
\end{tabular}
\end{center}
\end{table}

Even if we compare in terms of all the actions in a video, instead of only the actions belonging to a summary, there are clear differences. The main benchmark datasets for action recognition in general-purpose videos \cite{jiang2014thumos, caba2015activitynet} have specific properties: a mean video duration between 100 and 200 seconds, the whole actions represent at least 30\% of the video, and less than 20\% of the videos contain more than one type of action. However, in soccer datasets \cite{giancola2018soccernet} the mean video duration is more than one hour long, the actions represent at most 20\% of the video and all the videos contain more than one type of action (see Table \ref{tab:datasets_comparison}). 

It is important to mention that even though there are soccer-related datasets for action spotting \cite{giancola2018soccernet, jiang2020soccerdb} and team statistics \cite{pappalardo2019public}, to the extent of our knowledge there are no benchmark datasets for full match soccer summarization.

\subsection{Video Content vs Event Data}
One of the main challenges for the broadcasting companies in the recent years is the speed at which the customers expect the content to be available. The broadcasters need to provide summaries as soon as the match is finished. Despite this need, broadcasting companies usually do not rely on automatic algorithms to generate these summaries, instead they mainly rely on human editors who exploit event-metadata before video content since processing video content is very time consuming and browsing it to retrieve information is not easy. 

Automatized algorithms on visual content might help to reduce the work of human operators, but the amount of information to process is huge. A soccer match duration is at least 90 minutes, at a rate of 25 frames per second it corresponds to 135K frames of at least 112x112 pixels (C3D \cite{tran2015learning} input size). In addition, there are still some important challenges for these algorithms in terms of video content. The homogeneity of the field and the visual similarity between different types of actions make more difficult the discrimination of relevant and non-relevant information. The quick movement of the players around the entire field, the low resolution and the occlusions make their detection and tracking difficult. The subjectivity in broadcasting editorial decisions can lead a replay or a person-related closeup to miss information of the course of the match which can cause a disruption in time continuity of the video.

On the other hand, there exist event-metadata provided by companies like Prozone, GeniusSports, Opta, WyScout, and many others. These companies have human observers in the stadiums to collect on live the events happening on the field. The metadata describe all the match-related events like pass, shot, out, head, free-kick, corner, foul, cards, etc. The number of events in a match is significantly smaller than the number of frames. In the case of soccer, a match has about 1500 events overall, each event is represented by few values like the type, location and the distance to the goal. 

\subsection{Event and Action}
\begin{figure}[!ht]
	\begin{center}
		\includegraphics[width=0.4\textwidth]{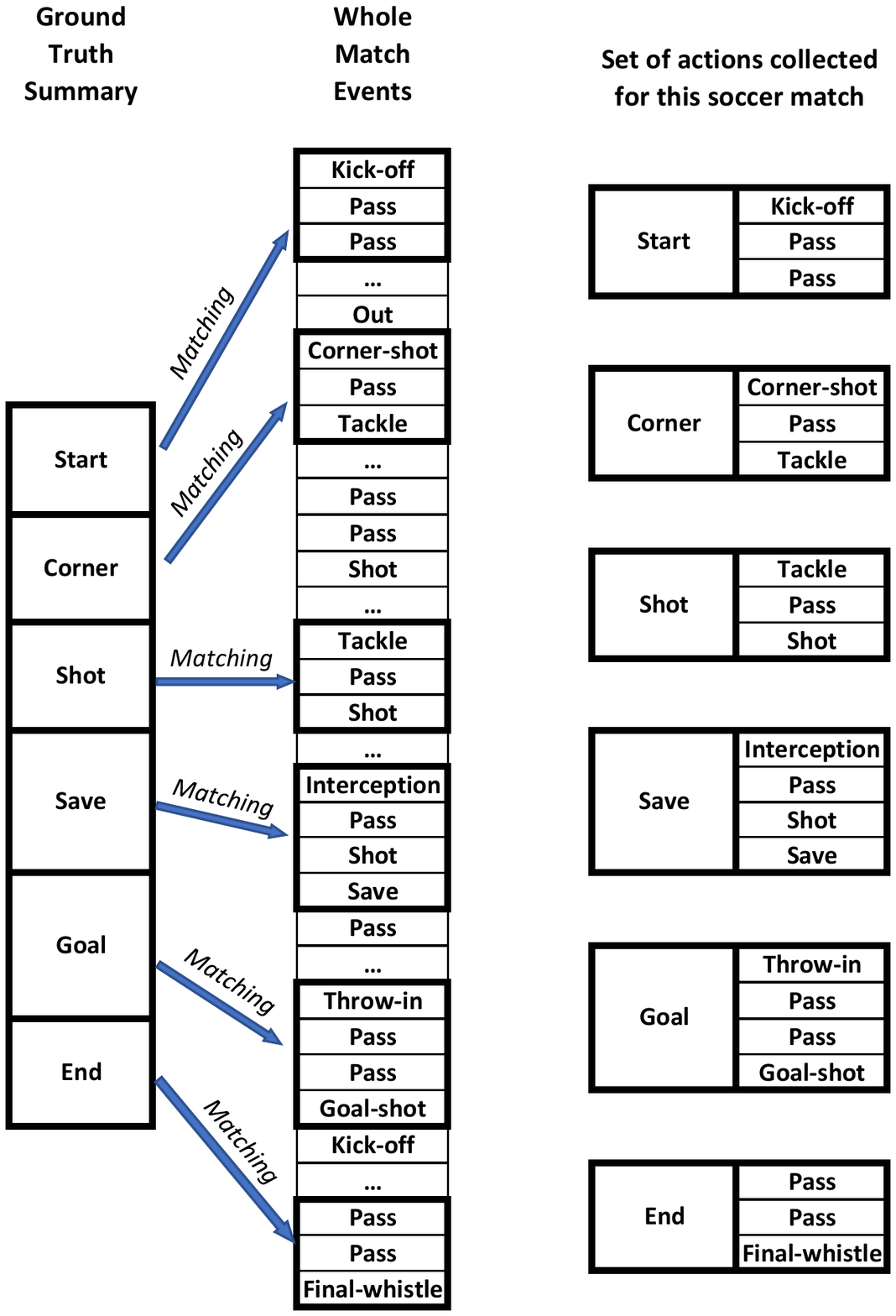}
	\end{center}
	\caption{Example of the event representation of a whole soccer match, displayed in the middle column. Events are all the atomic activities happening on the field during the given match. The first column represents the ground-truth summary of the given whole match. An action is a set of consecutive events that belongs to a summary.}
	\label{fig:events_actions}
\end{figure}
\label{sec:event_and_action}
In the multimedia community, the concept of event is generally vague and overlaps with the concept of action and activity. Chen et al. \cite{chen2014actionness} define the concept of action and actionness with 4 aspects that define an action: an agent, an intention, a bodily movement, and a side-effect. Dai et al. \cite{dai2017temporal} define an activity as a set of events or actions, with a beginning and an ending time. Sigurdsson et al. \cite{sigurdsson2017actions} argue that temporal boundaries in activities are ambiguous.

In our work, we define the concept of \textit{event} as any of the atomic activities happening on the field such as \textit{Pass, Tackle, Out, Goal-shot, etc}. This definition is very similar to the one described by Giancola et al. \cite{giancola2018soccernet}, where an event is anchored in a single time instance. Thus, an \textit{action} is a continuous set of consecutive \textit{events}. This way of defining actions allows to disambiguate their temporal boundaries. Furthermore, the events correspond to the metadata collected by Opta, Wyscout, and other companies. In the context of soccer, it is unclear when a given action such as \textit{scoring a goal} begins and ends. For this reason, we use as reference the different video clips of the summary videos. However, these video clips do not contain only events. For instance, the clip containing the event \textit{Goal-shot} might also show some events leading to the goal, then the celebration of the players, the reaction of the crowd or of the coaches and replays.

In our context, we want to reduce as much as possible the subjectivity, later to be added by editorial decisions, hence we consider only sports-related events and define an \textit{action} as the set of consecutive \textit{events} that might belong to a summary video clip. To give a more concrete example, we illustrate how we define events and actions in Figure \ref{fig:events_actions}. On the left, we show the \textit{actions} of the summary of the given match. Then, we match these \textit{actions} with the event sequences which correspond the best in the whole soccer match. We thereby obtain a set of event sequences that belong to a true summary. As aforementioned \textit{events} are all the atomic activities happening on the field during this match.



\section{Our Approach}
\label{sec:our_approach}
Most of state-of-the-art methods aim at summarizing a video in only one step, trying to identify the key frames that increase the diversity of the resulting summary. However, that approach might lead to two limitations in the summarization of sport matches. First, the diversity is not the main objective for a sports summary, for instance in soccer, the summary contains all the goals of the match and they all might be very similar. Second, the summary of a soccer match is not just sparse key frames spread along the video, each clip of the summary must represent an action. Therefore, similarly to \cite{sanabria2019deep, sanabria2021hierarchical} we split the summarization process in two tasks, first detect all the actions of the match that could be selected to be in the summary, and then decide which of these actions indeed belong to the summary.

Our approach is composed of three stages. The \textit{Generation of Action Proposals} stage (Section \ref{sec:proposals_geneation}) gets as input the event data of the match and uses Multiple Instance Learning for sequential data  to detect all the action proposals of the match. Then the \textit{Multimodal Summarization} stage (Section  \ref{sec:multimodal_summarization}) gets the event data and audio features of the action proposals and using a hierarchical multimodal attention model it decides which of these action proposals indeed belong to the summary. And finally, the \textit{Multiple Summaries Generation} stage (Section \ref{sec:multi_summaries}) uses a ranking distribution in order to provide several summary options of the same match to the editor.
\subsection{Generation of Action Proposals}
\label{sec:proposals_geneation}
The goal of the first stage of our method is to identify the action proposals of the match, that is to say consecutive relevant events. Such groups of events are considered as proposals if they are parts of the match that might belong to the summary.

\subsubsection{Similarity of inter-categorical actions: Multiple Instance Learning}
\label{sec:MIL}

\begin{figure*}
	\begin{center}
		\subfloat[Inter-categorical Similarity]{%
			\label{fig:mil_inter}
			\includegraphics[keepaspectratio, width=0.75\textwidth]{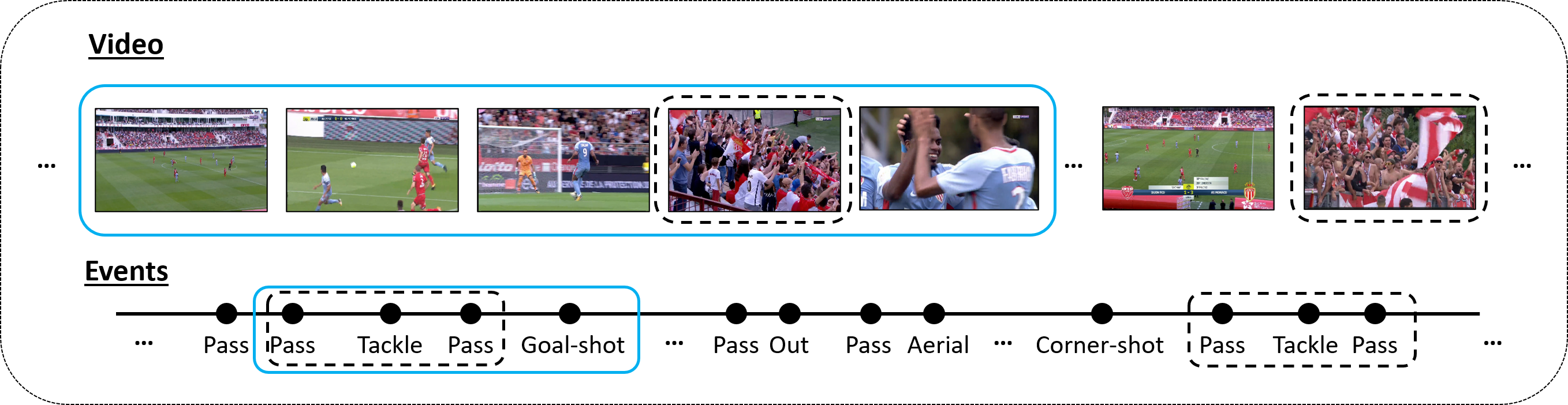}}
		\hfill
		\subfloat[Intra-categorical Diversity]{
			\label{fig:mil_intra}
			\includegraphics[keepaspectratio, width=0.8\textwidth]{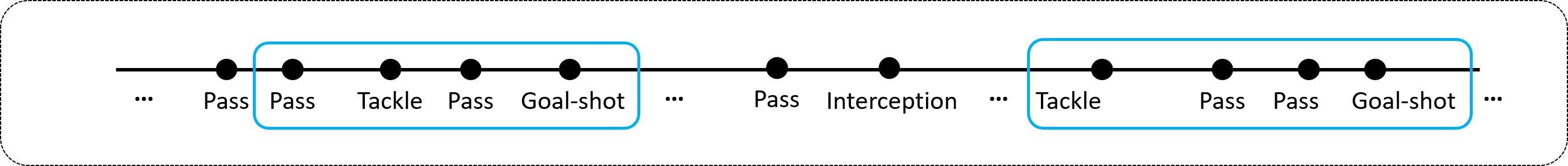}}
		\hfill
		\caption{Examples of inter-categorical similarity and intra-categorical diversity. (a) Example of inter-categorical similarity of actions. In the bottom part there are video frames that represent clips of video. In the bottom, the black dots are the events of the match. Blue line represents a goal action. The dashed line indicates that similar parts of the match can be both inside an action and outside an action. (b) Example of intra-categorical diversity of actions. Blue line represents a goal action. Two goal actions might be formed by different sequences of events.}
	\end{center}
\end{figure*}

The idea of extracting pieces of the input as proposals to then in a second stage decide which of these proposals are indeed classified as positive has been widely used in object detection \cite{ren2015faster, girshick2015fast, kong2016hypernet, li2018high, tang2018weakly,  shih2019real} and action detection \cite{buch2017sst, gao2017turn, xu2017r, dai2017temporal, guo2018fully}. The main goal of the proposals extraction is to filter as much as possible the relevant and non-relevant information by identifying the negative parts of the sample (i.e.,background in the case of object detection and non-action in the case of action detection), in order to be discarded for the following classification stage.

However, in sport matches there is a high similarity between positive (i.e.,action) and negative (i.e.,not action) samples. In the case of soccer, for instance the sequence $\{Pass, Tackle, Pass\}$ can be the beginning of a goal action but the same sequence can belong to some section of the match where nothing relevant is happening (see Figure \ref{fig:mil_inter}). This inter-categorical similarity is not the only issue we have to face, since simultaneously our system should be able to deal with a high intra-category diversity where two instances of the same action can only partially match when considering their event sequences (see Figure \ref{fig:mil_intra}). For these reasons, we believe that in the context of soccer matches, a Multiple Instance Learning (MIL) approach is more suitable than a traditional learning method. 

MIL paradigm was first described by Dietterich et al. \cite{dietterich1997solving} to predict drug activity. Then it was introduced in many other fields like object tracking \cite{babenko2010robust}, object detection \cite{tang2017multiple, cinbis2016weakly} and image tagging \cite{rahman2019deep0tag}. This paradigm is very popular in medical images \cite{mercan2017multi, sudharshan2019multiple, quellec2017multiple}, where an entire image of the organ or tissue is labeled as malign but only a small portion of it is actually malign.

In the traditional supervised classification problem, a model predicts a target value for a given instance. In the MIL paradigm, instances are grouped in labeled \textit{bags} (a \textit{bag} is hence a set of instances), without the need that all the instances of each bag have individual labels. In the binary classification case, a bag is labeled positive if it has at least one positive instance; on the other hand, a bag is labeled negative if all its instances are negative.

\begin{equation}
Y =
\begin{cases}
 0, & \text{iff $\sum_{k} y_k = 0$,} \\
1, & \text{otherwise} \\
\end{cases}
\end{equation}

For our action proposals problem in soccer matches, we follow the event and bag representation proposed in \cite{sanabria2021hierarchical}. A match is a sequence of events \{$e_1, e_2, ..., e_F$\} representing all the events occurring on the field. We denote a bag $b_n = \{ e_1^{b_n} e_2^{b_n}, ..., e_K^{b_n}\}$ as a set of consecutive events. $X=\{x_{e_1},x_{e_2},...,x_{e_F}\}$ represents the set of instances, where $x_{e_f}$ is the feature vector characterizing the $f$-th event of the match. We assume that $K$ could vary for different bags. The difference between action and bag is that an action is a set of consecutive events that might belong to the summary, while a bag is a set of any consecutive events.

\subsubsection{Multiple Instance Learning for Sequential Data}

MIL was originally proposed as a general and abstract learning paradigm and, as such, it does not require or involve any feature extraction process. However, recent works consider MIL into an end-to-end trained model, within the process of learning features by using a fully-connected neural network \cite{wang2018revisiting,ilse2018attention}. According to the assumptions of the paradigm, a MIL model must be permutation-invariant where the instances exhibit neither dependency nor ordering within a bag. However, these characteristics do not necessarily apply in soccer actions since the order of the events can clearly impact the meaning or interest of an action. We believe that fully-connected layers as proposed by previous state-of-the-art works are not completely suitable to capture this sequentiality. For this reason, we use an LSTM network followed by a MIL Pooling to get the bag representation. Our LSTM MIL Pooling method is similar to the one proposed by Janakiraman et al. \cite{janakiraman2018explaining} but instead of using the MIL aggregation at the prediction level, our method performs the aggregation at the feature level.

In our LSTM MIL Pooling method, a bag is a sequence of event feature vectors. This sequence is the input of an LSTM with hidden state defined by:

\begin{equation}
h_t = LSTM(h_{t-1}, x_{e_f}^M)
\end{equation}

\begin{figure}
\begin{center}
 \includegraphics[width=7cm]{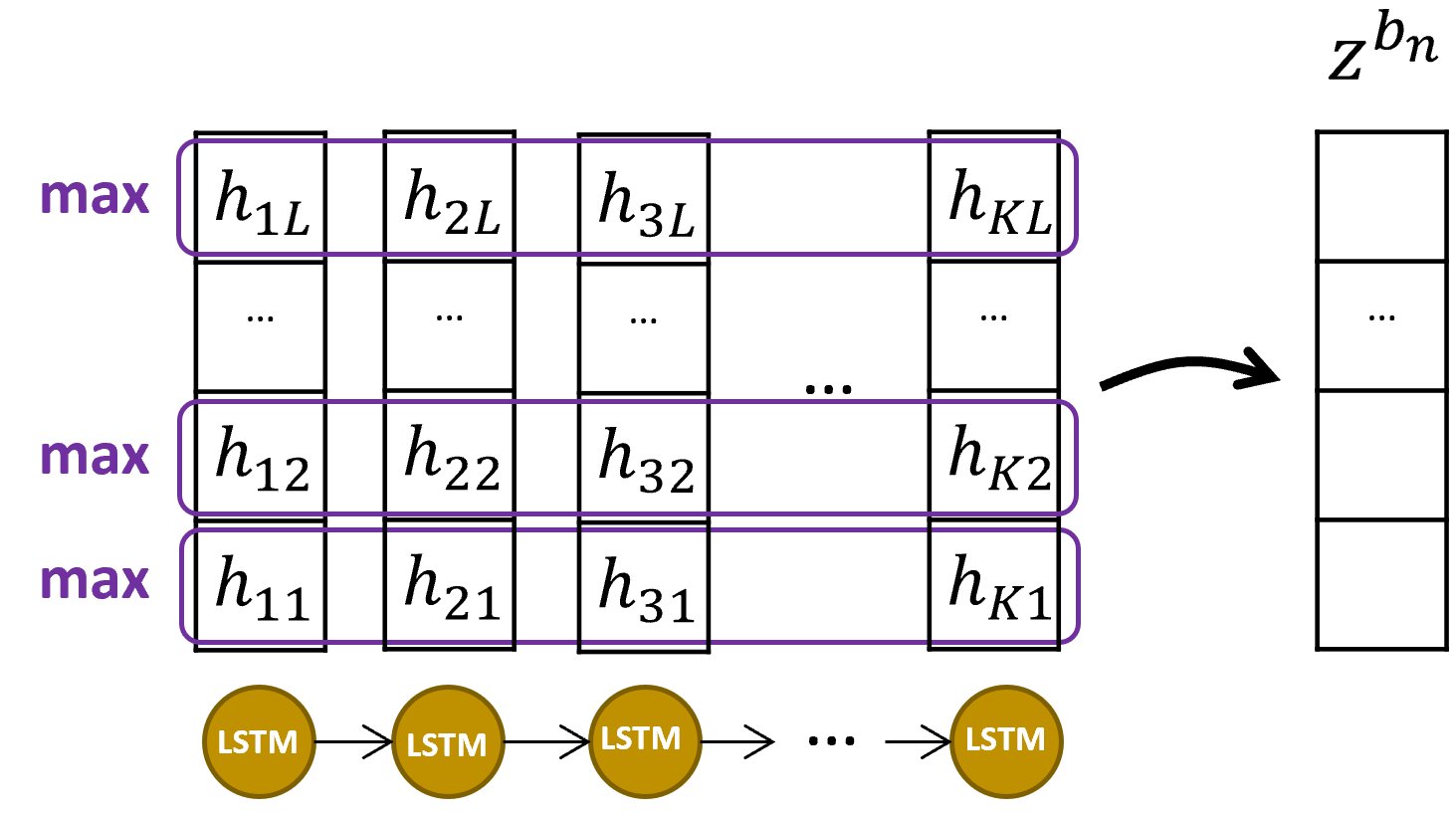}
 \end{center}
 \caption{LSTM MIL Pooling schema for a bag $b_n$. $h_kL$ is the hidden state of size $L$ for event k.}
\label{fig:mil}
\end{figure}

where $x_{e_f}^M$ is the metadata feature vector of event $e_f$ and $LSTM(h_{t-1}, x_{e_f}^M)$ represents an LSTM function of hidden state $h_{t-1}$ and input vector $x_{e_f}^M$. This recurrent network learns an embedding for each event preserving the sequential dependency between events.

Let $H^{b_n} = \{h_1,...,h_k\} $ be the $K$ embeddings of the $K$ events from bag $b_n$. Each $h_k$ embedding is of size $L$. Then the MIL Pooling step to learn the final bag representation $z^{b_n}$ is defined in Eq.(\ref{eq:mil_pooling}), where $z^{b_n}$ is a feature vector of size $L$, and is obtained from getting the maximum of each position $l$ across all the $K$ event embeddings of the bag (see Figure \ref{fig:mil}).

\begin{equation}
\forall_{l=1,...,L} : z_l^{b_n} = \max_{k=1,...,K} {h_{kl}}
\label{eq:mil_pooling}
\end{equation}

This representation $z^{b_n}$ is the input of a single sigmoid neuron which provides the score $O_{b_n}$, a value between $0$ and $1$, for the bag $b_n$ to be an action proposal or not.

\subsubsection{Action Proposals}
\label{sub_sec:action_proposals}

It is important to notice that we do not have access to the ground-truth actions, then the bags are created in a class-agnostic way. We use a sliding window across the whole match events, with a stride such that there is an overlap between two consecutive windows (Bottom of Figure \ref{fig:approach_diagram} shows an example of the sliding window). 

The overlap on the sliding window leads some events to belong to more than one bag and thus obtaining several scores. In order to define a score per event instead of per bag, we merge all the possible scores associated to the given event. The score $S_{e_f}$ for the event $e_f$ is then given by the Log-Sum-Exp (LSE) used in \cite{sanabria2019deep}, the function is defined in Eq. (\ref{eq:lse_function}). The LSE is a smooth version and convex approximation of the max function. The hyperparameter r controls the smoothness of approximation \cite{wang2018revisiting}.

\begin{equation}
S_{e_f} = r^{-1} \cdot log \Bigg[ \frac{1}{|\{b_n \mid e_f \in  b_n\}|} \sum_{b_n \mid  e_f \in  b_n} r \cdot O_{b_n} \Bigg]
\label{eq:lse_function}
\end{equation}

After obtaining the score per event, we find a threshold using the validation set and we consider as positive all the events with a score higher or equal than the threshold. Thus an action proposal $a_p$ is a set of positive consecutive events (see Figure \ref{fig:approach_diagram}). 

We manually decide the end of an action only in one specific case, when one of the events inside the action is a goal-shot. For instance, we found that $\{e_7, e_8, e_9, e_{10}, e_{11}\}$ is a set of positive consecutive events and both $e_6$ and $e_{12}$ are negative events. If one of these events let us say $e_9$, is goal-shot, then we would define two different actions, $\{e_7, e_8, e_9\}$ and $\{e_{10}, e_{10}\}$.

\subsection{Multimodal Summarization}
\label{sec:multimodal_summarization}

In Section \ref{sec:proposals_geneation} we have described how we use Multiple Instance Learning to obtain a score per event and how we have defined that an action proposal $a_p$ is a set of positive consecutive events. Now the goal of the second stage of our approach is to define which of these proposals indeed belong to the summary.

In broadcasting companies, the human editors rely on different sources of information to decide which actions are part of the summary. Audio is one of these sources since the excitement in the commentators' voice and the crowd cheering play a very important role to decide the importance of an action. 

Therefore, we propose a hierarchical multimodal attention mechanism (HMA) that in the first stage learns the importance of each modality (audio and metadata) at the event level and in the second stage learns the importance of each event inside the action proposal. We believe that the multimodal attention at the event level instead of the action level is very useful in sport actions since the importance of each modality can significantly vary from one event to the other inside the same action. 


Let us describe the two stages of our method. In the first stage of the hierarchy, the multimodal representation vector per event is given by a weighted average: 

\begin{equation}
c_i = \lambda_i^M h_i^M + \lambda_i^A h_i^A
\end{equation}

The weights of each modality $\{\lambda_i^M, \lambda_i^A\}$ are determined by an attention layer that shares the parameters $W$ across time-steps:

\begin{equation}
\lambda^{\{M, A\}}_i = softmax(tanh(W(h_i^{\{M, A\}})))
\end{equation}

After obtaining a multimodal representation per event $c_i$, there is an attention layer that learns the importance of each event inside the action proposal, resulting in the weight $\beta^c_i$:

\begin{equation}
\beta^c_i = softmax(tanh(h_i^c))
\end{equation}

Thus, the representation vector per action proposal is given by a weighted average: 
\begin{equation}
{d_{a_p}} = \sum_{i=1}^{L_p} \beta_i^c h_i^c
\end{equation}
where $L_p$ is the length (number of events) of action $a_p$.
Finally, each of this $d_{a_p}$ action representation is given to a sigmoid neuron which outputs a value between 0 and 1, that indicates the likelihood of the action $a_p$ to be included in the summary.


\subsection{Multiple Summaries Generation}
\label{sec:multi_summaries}

In Section \ref{sec:proposals_geneation} we have described how we use Multiple Instance Learning to detect action proposals to then in Section \ref{sec:multimodal_summarization} exploit the metadata and audio features of the match to define which of these proposals indeed belong to the summary. However, as mentioned before, the subjectivity is a big challenge in sports summaries. As a solution to this subjectivity, we design a strategy to propose candidate summaries giving the editors the chance to choose the best option according to their needs. 

On the other hand, the wide use of social networks has changed the way new generations consume the content, users prefer shorter clips available in different online platforms. Then, during the creation of the summaries, the editors of broadcast companies are generally constrained by the length of the resulting video summary. 

The natural way of creating a summary in a time-constraint context is to rank the actions by importance and then adding one by one until the summary’s specified time-budget is filled. To give more freedom to the editor we would like to propose several rankings of the same match so they can generate several candidate summaries. 

\subsubsection{Ranking Distribution}

One way to generate several rankings is to sample from a ranking distribution. A classic model for such distribution is the Plackett–Luce model, which was proposed by Plackett \cite{plackett1975analysis} to predict the ranks of horses in gambling.

The Plackett-Luce model is parameterized by a vector $\theta = (\theta_1, \theta_2, ...,\theta_P) \in \mathbb{R}_{+}^{P}$. Each $\theta_p$ can be interpreted as the \textit{importance} of the option $p$, with higher values indicating that an item is more likely to be selected \cite{el2018ranking}. 

Consider a fixed set $A=\{a_1, ..., a_P\}$ of P action proposals. We identify a ranking over $A$ with a permutation $\pi \in \mathbb{S}_P$, where $\mathbb{S}_P$ denotes the collection of permutations on $[P] = \{1,...,P\}$. Thus, each $\pi$ is a bijective mapping $[P] \rightarrow [P]$. The probability assigned by the Plackett-Luce model to a ranking represented by a permutation $\pi \in \mathbb{S}_P$ is given by

\begin{equation}
\mathbb{P}_{\theta} (\pi) = \prod_{i=1}^{P} \frac{\theta_{\pi^{-1}(p)}}{\theta_{\pi^{-1}(p)} + \theta_{\pi^{-1}(p+1)} + ... + \theta_{\pi^{-1}(P)}}
\label{eq:pl}
\end{equation}

A ranking of $P$ items can be viewed as a sequence of independent choices: first choosing the top-ranked item from all items, then choosing the second-ranked item from the remaining items and so on. In each step, the probability of an item to be chosen next is proportional to its \textit{importance}. Consequently, items with a higher \textit{importance} tend to occupy higher positions. In particular, the most probable ranking is simply obtained by sorting the items in decreasing order of their \textit{importance} (like the previously explained natural way of creating a summary).

\subsubsection{Ranking actions}

The hierarchical multimodal approach explained in Section \ref{sec:multimodal_summarization} gives as output a value between 0 and 1 that can be interpreted as the importance of each action. Therefore, the values of $\theta$ are determined by the output of the last layer of the multimodal summarization model.

To be more specific, if there are $P$ actions detected in a match and therefore a $\theta_p$ value for each of these actions, each of the candidate summaries is built by sampling $pl$ from the Plackett-Luce distribution with parameter $\theta$. A way to sample from the Plackett-Luce distribution is to sort Gumbel perturbed log-scores \cite{grover2019stochastic} as described in Equation \ref{eq:pl_sampling}:

\begin{equation}
pl = argsort(\log \theta + g)
\label{eq:pl_sampling}
\end{equation}

where $g = Gumbel(0, \sigma)$. 

Let us consider a specific example. First the Generation of Action Proposals stage provides $P$ actions in a specific match, after the HMA outputs a score $d_{a_p}$ for each of these actions. These scores are considered as the importance $\theta$ for the Plackett-Luce distribution. Then to generate a candidate summary, we sample a ranking from the distribution $pl$. The output of this sampling is a list of size $P$ that indicates the position of each action, where the higher the ranking the higher the importance.  The number of actions chosen to be part of the candidate summary is determined by the ground-truth summary length, the actions are chosen until the candidate summary has a duration that is very close to the ground-truth video summary duration. Finally the candidate summary video contains the video clips of     each chosen action, in the order they occur in the match. 

For instance, in Figure \ref{fig:approach_diagram} the first sampling $pl_1$ generates a ranking where the action $a_3$ is the most important followed by $a_{10}, a_5, a_1 a_9$ and so on. And the candidate summary generated from this ranking contains only the actions $a_3, a_{10}, a_5$ and $a_1$ since these 4 actions already comply the time-constraint. 

\section{Experiments}
\label{sec:experiments}

In this section we introduce the experimental setting, describing the way we created the different datasets, the multimodal features and the metrics.  

\subsection{Summary Video Datasets}
There are widely used datasets focusing on sports such as UCF Sports \cite{rodriguez2008action}, Sports1M \cite{karpathy2014large}, Olympics Sports \cite{niebles2010modeling} but they serve the objective of video classification and are not suitable to evaluate video summarization. SoccerNet \cite{giancola2018soccernet, deliege2021soccernet} and SoccerDB \cite{jiang2020soccerdb} are datasets dedicated to soccer videos. SoccerNet is a benchmark for action spotting which only provides one-second annotation for different type of events. SoccerDB provides data for object detection, action recognition and temporal action localization, but they do not provide the information for entire matches, and they assume a clip is a highlight if there is a playback in it.


Based on the assumption that each clip inside a summary follows a logical time sequence decided by a human editor where a main event (e.g. goal-shot, card, free-kick) is shown along with its context (e.g. the foul that led to the card), we need a specific dataset where summary actions are present with their context. To the extent of our knowledge, there are no publicly available datasets for soccer video summarization. We therefore create our own. The first video dataset consists of 100 matches from the 2019-2020 season of the English Premier League. We take the exact videos broadcast on TV. The only ground-truth available for these matches are the 100 video summaries created by professional broadcasters. The duration of these video summaries varies from 110 to 270 seconds.

Pappalardo et al. \cite{pappalardo2019playerank} released a set of soccer-logs collected by Wyscout, containing all the spatio-temporal events that occur during all matches of an entire season of seven competitions. As a second video dataset, we choose the World Cup 2018 data since it is the competition for which we could find on Youtube the most number of matches and their corresponding summary video. We use 56 matches from the 64, since we do not consider the matches with extra times and the matches for which both match and summary videos are not available. This second video dataset will be later used for evaluating transfer learning and generalization capability of our method.

\subsection{Action Vocabulary}
\label{sec:act_dataset}

A summary video is composed of multiple clips extracted from the whole match video. However, there is no formal definition of what is an action that should belong to a summary, even formally defining any (generic) action is not simple. The only information we can have access to for the whole match data are beginning and end timestamps of each event happening during the match. As explained in Sec.~\ref{sec:event_and_action}, in our context we define relevant actions as all the sequences of events which can be found in any summary of the training set. 

\begin{figure}[!h]
\begin{center}
 \includegraphics[width=0.5\textwidth]{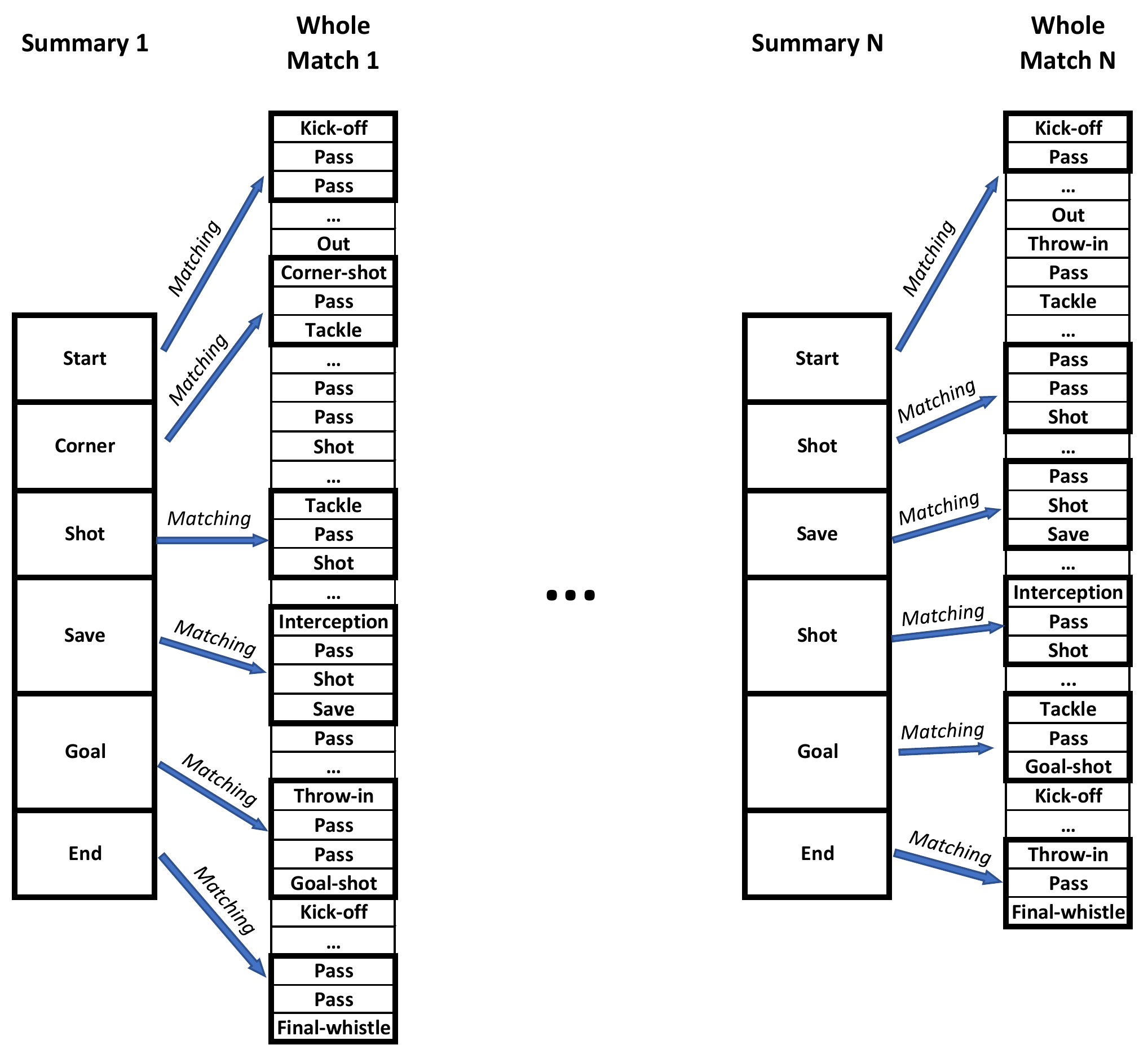}
 \end{center}
 \caption{Example of collecting actions, i.e.,sequences of events from all training summaries.}
\label{fig:all_events}
\end{figure}

\begin{figure}[!h]
\begin{center}
 \includegraphics[width=0.5\textwidth]{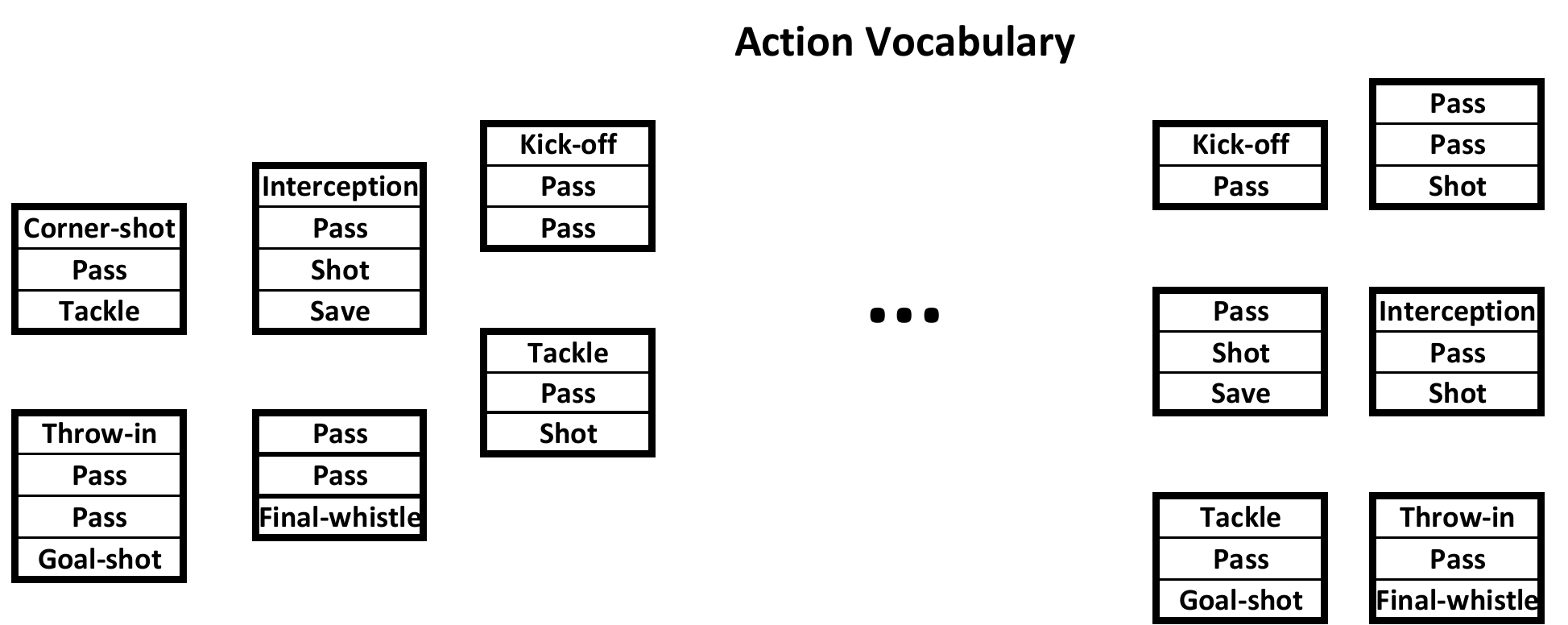}
 \end{center}
 \caption{All the actions collected from summaries in the training set and the corresponding sequences of events in all the matches in the training set.}
\label{fig:action_vocabulary}
\end{figure}

As displayed on Fig.~\ref{fig:all_events}, each summary provides a set of actions thus collecting the corresponding sequences of events from the corresponding whole matches. This process results in a vocabulary of actions that is going to be the basis for detecting potential summary actions (see Figure.\ref{fig:action_vocabulary}). 

\begin{figure}[!h]
	\begin{center}
		\includegraphics[width=0.35\textwidth]{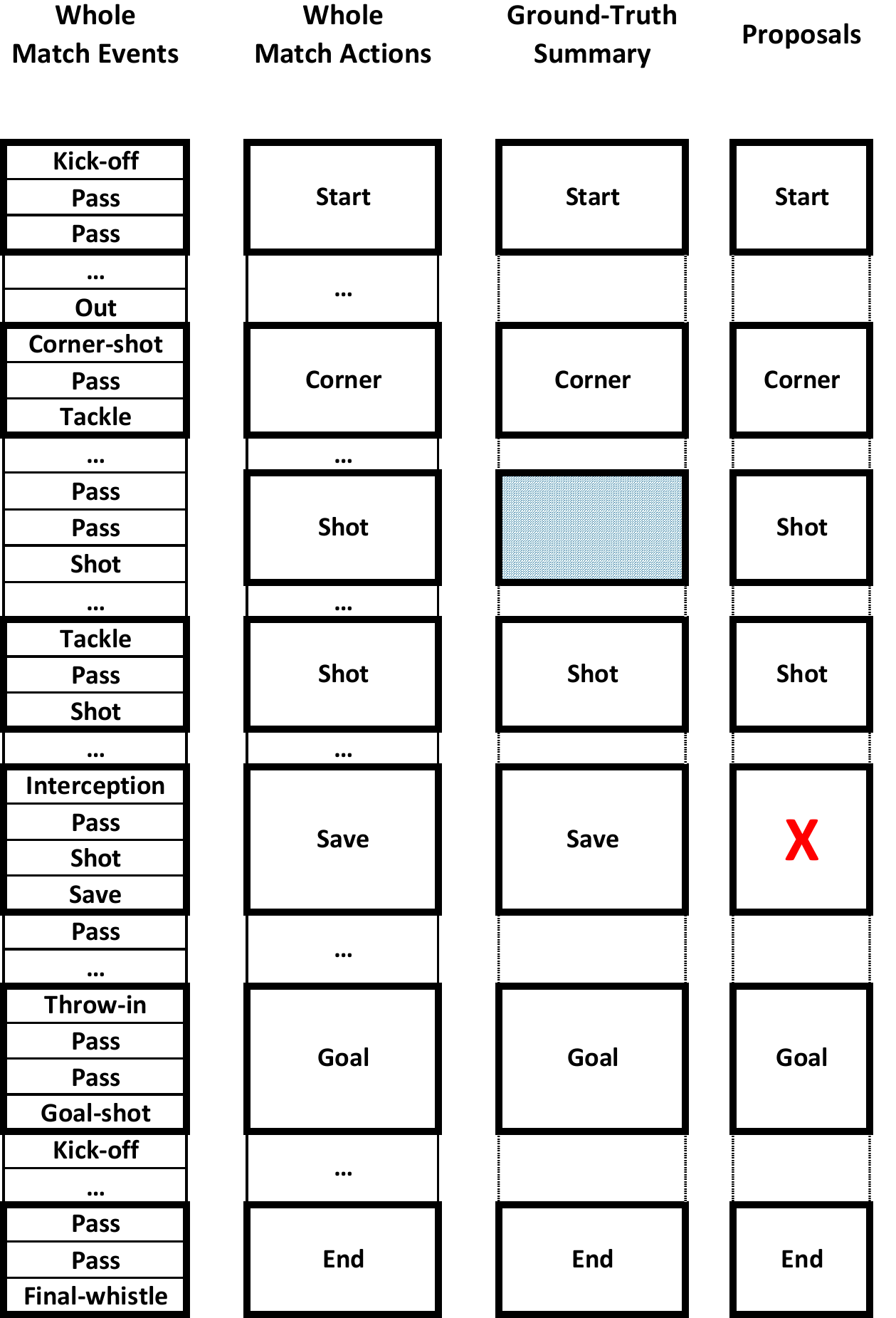}
	\end{center}
	\caption{Example of the event representation of a whole soccer match, displayed in the first column. Events are all the atomic activities happening on the field during the given match. An action is a set of consecutive events that might belong to a summary. The third column represents the ground-truth summary of the given whole match, the blue cell indicates the action does not belong to the summary of the given match. A proposal is a set of consecutive positive events predicted by our Action Proposal Generation stage. The fourth column describes thus the set of proposed actions to be possibly in the final summary.  The red cross \textcolor{Red}{X} indicates that the action was not predicted as proposal.}
	\label{fig:event_actions_example}
\end{figure}

Let us give a more concrete example, we illustrate the process in the first two columns of Figure \ref{fig:event_actions_example}. In the first column on the left, we present an example of the event representation of a given soccer match. As aforementioned \textit{events} are all the atomic activities happening on the field during this match. In the second column, we show the \textit{actions} of the given match, thus the event sequences (corresponding to these actions) belong to the summary of this match or to the summary of any other match in the training set. This explains also the diagonal stripes area which represent an action, here the sequence of events \textit{\{Pass, Pass, Shot\}}, which is not part of the final Ground Truth Summary of that given match (produced by a human operator) but the exact same sequence is part of the Ground Truth Summary of another match in the training set.

It is thus important to note that following this definition, many events do not belong to any action. For instance, in the first column on the left of the Figure \ref{fig:event_actions_example}, the event \textit{Out} happening before the event \textit{Corner-shot} or the event \textit{Kick-off} right after the event \textit{Goal-shot}, are not part of any action since they do not belong to an event sequence of the current summary as it can be seen in the third column corresponding to the Ground Truth Summary for that match, but they do not belong either to any other sequence from any other summary in the training set. 

The final step in the creation of the action proposal dataset is to label as action any sequence of events that is identical to any of the actions vocabulary.

A more detailed explanation is given by a joint analysis of the columns \textit{Whole Match Actions} and \textit{Ground Truth Summary} from Figure \ref{fig:event_actions_example}. All the actions of the ground-truth summary are part of the action proposals of the match. And even though the sequence of events \textit{\{pass, pass, shot\}} is not in the ground-truth summary of the corresponding match (Blue cell), it is still considered as a potential action proposal for this match because this sequence of events was found in the ground-truth summary of another match from the training set. It is important to notice that to find the action proposals, we only look for actions present in matches from the training set and not in any match from the test set or the validation set.\\

An intuitive approach in such a context is to use Template Matching to ``predict'' in the validation or test data which sequences of events should be considered as actions. However with this approach, we would be only able to recognize sequences of events that have already been seen in the training set and only exactly these sequences. Furthermore any action which is missed at this stage will never be retrieved.

Instead we propose to use weak supervision and Multiple Instance Learning (MIL) principles from the summary actions to be able to capture all possible variations in these actions.

\textbf{Training}. All the sequence of events matching with the action proposals are considered as positive bags. To obtain the negative (no-action) bags, we randomly pool sequence of events from the parts of the match where there are not events labeled as action. It is important to recall that not all the events of the match belong to an action, as it is shown in Figure \ref{fig:event_actions_example}, for the Out event before the Corner action or the Kick-off event after the Goal action. 

The length of the pooled sequence of events varies from 4 to the maximum action proposal size in the training set. Finally, we randomly choose as many negative samples as the positive samples. We then train our sequential MIL model on all these sequences of events (as defined in Section \ref{sec:MIL}).

\textbf{Testing}. As previously mentioned, in a real-life scenario we do not have access to the ground truth intervals in the test phase, the bags are created in a class-agnostic way. We use a sliding window across the whole match events, with a stride such as there is an overlap between two consecutive windows, and we predict on these bags.

For the Generation of Action Proposal stage, we only use event data features since we consider the audio features are not relevant to distinguish between action and no action. On the other hand, we consider that the use of multiple modalities is very important to decide if an action is relevant. For this reason, we use event data and audio features in the Multimodal Summarization stage. 

\subsection{How to evaluate the Generation of Multiple Summaries}
As mentioned before, there is not a unique and perfect ground-truth summary for a match due to subjectivity generally present in this task. A clear example is given by the goal-attempt actions, it is evident that the common actions between all the possible summaries are the goals of the match, but how can we assure that the goal-attempt of 4 minutes before the goal is more or less relevant in the summary than the one 6 minutes before? Sometimes the editorial reason to add some goal-attempts is just to show the persistence or the lead of a team. Then we can argue that the evaluation is not really fair when a non-detected goal-attempt is counted as false negative, and another very similar goal-attempt is counted as false positive.

Therefore, instead of evaluating the multiple generated summaries only in terms of time intervals, we consider that a ground-truth action is correctly detected if the predicted action meets two conditions:
\begin{enumerate}
  \item The action type is the same.
  \item It occurs in the time interval between the previous and the following ground-truth action.
\end{enumerate}

For instance, to check if the third action of the ground-truth summary is detected, we check if there is a predicted action of the same type between the time interval $[time_a, time_b]$, where $time_a$ is the end time of the second ground-truth action and $time_b$ is the start time of the fourth ground-truth action.

We defined 10 types of actions $T$ which cover all the possible actions of a summary: free-kick, corner, foul, shot, save, referee decision with video assistance (VAR), goal, end period, start period, other. An action is labeled with a certain type $T$ if at least one of the events of the action is type $T$.

\section{Results}
\label{sec:results}

For a fair comparison we use 10-fold-cross-validation. Each fold has $80\%$, $10\%$ and $10\%$ of the matches for train, validation and test set, respectively. For all the comparison experiments, we replicate the models from the source paper using Keras library and choose the parameters with the highest classification performance on our validation dataset

\subsection{Generation of Action Proposals}

In order to choose the best method to detect the actions, we compare with three different methods: SST \cite{buch2017sst}, MI-Net \cite{wang2018revisiting} and MI-Net Attention \cite{ilse2018attention}. 

As we previously mentioned, the idea of generating proposals has been tackled by several approaches. We chose SST \cite{buch2017sst}, which was created to generate temporal action proposals for action detection in untrimmed video sequences. This method is a RNN-based architecture that at each time step produces confidence scores of different action sizes ending at this time step. Instead of using video features as it was originally proposed, we use our event data features as input. We implement an LSTM network with 16 neurons, the proposal sizes are $\{2, 3, 4, 5, 6, 7, 8, 9, 10, 11, 12\}$, it is trained with binary cross-entropy loss and Adam optimizer. In training phase, an output interval is considered positive if at least 50\% of it belongs to a positive action from the \textit{Actions Dataset}.

In terms of Multiple Instance Learning, we chose two neural networks based approaches. MI-Net is composed of fully connected layers to extract a representation per sample and then it gets a score per bag using a pooling layer with max operator over all the samples of the bag. MI-Net Attention replaces the max-pooling layer by an attention mechanism that learns the importance of each sample of the bag. Our LSTM MIL Pooling stage uses a recurrent neural network instead of fully connected layers to then get a score per bag using a max-pooling layer. 

The number of neurons for the fully connected layer of the MI-Net approaches are 32. The attention layer has 8 neurons. We use a stochastic gradient descent optimizer with a nestrov momentum 0.9, a weight decay 0.005 and initial learning rate 0.0005. The code is based on the authors implementation \cite{wang2018revisiting_github}.

LSTM MIL Pooling network has 16 neurons. We use Adam optimizer with 0.9 and 0.999 as the exponential decay rate for the 1st and 2nd moment estimates, learning rate 0.001, binary cross-entropy as loss function and a batch size of 32 bags.

For the evaluation phase, we consider an action was correctly classified if at least 50\% of the predicted action belongs to a positive action from the \textit{Actions Dataset} (Section \ref{sec:act_dataset}). 

Since the objective of the \textit{Generation of Action Proposals} stage is to detect all the possible actions of the match, we want to obtain the least false negatives rate possible. For this reason, we use the F2-score obtained in the validation set to choose the best threshold and best epoch per fold, as this score weights the recall twice as important as the precision. In order to use a sports terminology, from now on we will call \textit{Missing Actions} to the false negatives rate, representing all the ground-truth actions that were not detected.

\begin{table}[!h]
\begin{center}
\caption{Performance Comparison of Action Proposals Generation methods. The arrows indicate that the lower(higher) the value the better.}
\label{tab:mil}
\begin{tabular}{|c|c|c|}
\hline
\textbf{\thead{ Method}}
 & \textbf{Missing Actions $\downarrow$}  & \textbf{F2-score $\uparrow$}  \\
\hline
SST \cite{buch2017sst} & 22.01 & 54.25 \\
\hline
MI-Net \cite{wang2018revisiting} & 14.47 & 71.68\\
\hline
MI-Net Attention \cite{ilse2018attention} & 19.01 & 70.87\\
\hline
LSTM MIL Pooling & 8.82 & 73.73 \\
\hline
\end{tabular}
\end{center}
\end{table}

Table \ref{tab:mil} depicts the performance of the methods previously described. We will focus on the Missing Actions rate and F2-score. The Missing Actions rate is important since in this stage we will pay more attention on not losing any potential summary action of the match. And we consider the F2-score as a control metric since the easy solution to miss the least number of actions is to predict all the samples as positive, then the F2-score helps to verify the method has a reasonable Precision value.

MIL methods are clearly better at detecting the different actions of the match since SST performs at least 16\% worse than the rest of the methods. LSTM MIL Pooling outperforms all the other methods, it misses at least 5\% less actions and gets a F2-score at least 2\% higher compared with the second-best method (MI-Net).

The way of creating the Actions dataset proposed in Section \ref{sec:act_dataset}, where we take as reference the training set which represents 80\% of the matches, opens the question if it is really necessary a Generation of Action Proposals stage. One might argue that checking for the exact sequence of events of the training set on the testing set is enough to define the actions of the match. 

\begin{table}[!h]
\begin{center}
\caption{Detected actions before and after the Action Proposals Generation stage}
\label{tab:before_after_mil}
\begin{tabular}{|c|c|c|}
\hline
\textbf{\thead{ Method}}
 & \textbf{Missing Actions $\downarrow$}  & \textbf{F2-score $\uparrow$}  \\
\hline
Template Matching & 10.87 & 45.55 \\
\hline
LSTM MIL Pooling & 8.82 & 73.73\\
\hline
\end{tabular}
\end{center}
\end{table}

As ablation study of this stage, Table \ref{tab:before_after_mil} compares the results of before and after the Generation of Action Proposals stage. The scores of the first row of the table are obtained assuming there is no learning to detect the actions of the match, meaning that the actions are just the sequences of events that are identical to the reference actions (as described in Section \ref{sec:act_dataset}). 

After using LSTM-MIL Pooling we miss at least 2\% less actions and get 28\% more in F2-score. This can be explained analyzing the order of the events composing the actions. In the case of not having a Generation of Action Proposals stage, for instance if in the training set there are different actions formed from different combinations of the group of events $[interception, pass, pass, goal-shot]$ but there is not a single action with the exact sequence of events $\{pass, interception, pass, goal-shot\}$ found in the test set, which is the same set of events but in an order not found in any of the matches of the training set, then this action would be completely ignored by the Summarization stage. The same would happen with an action that contains the same group events but with some new event in the middle, e.g. $\{interception, pass, pass, tackle, goal-shot\}$. However, a method like LSTM-MIL Pooling might learn for instance that even if the same exact sequence of events is not present in the training set, a goal-shot event is always part of an action of the match. 

\subsection{Multimodal Summarization}
\label{sub_sec:summarization}

The input of our Multimodal Summarization stage are the actions described in Section \ref{sub_sec:action_proposals}. 

We compare our Hierarchical Multimodal Attention (HMA) model with two of the most widely adopted structures for multimodal attention, we call them \textit{One-Level Attention} and \textit{Two-Level Attention}. \textit{One-Level Attention} computes a vector per modality and then uses an attention model to learn the importance of each modality \cite{xu2017learning, caglayan2016multimodal}. \textit{Two-Level Attention} uses an additional attention model for each modality independently \cite{hori2017attention, li2020aspect, zhu2018msmo}. The schema of these two models are described in the Supplementary Material.

For the model parameters, we use 32 neurons for the LSTM of each modality, Adam optimizer with 0.9 and 0.999 as the exponential decay rate for the 1st and 2nd moment estimates, learning rate 0.001, binary cross-entropy as loss function and a batch size of 32 bags. Our model has 32 neurons in $h^M$ and $h^A$, and 16 neurons in $h^c$.

\begin{table}[!h]
\begin{center}
\caption{Performance comparison of Multimodal Attention methods.}
\label{tab:attention}
\setlength{\tabcolsep}{5pt}
\begin{tabular}{|c|c|c|c|}
\hline
\textbf{\thead{ Method}}
 & \textbf{Missing Actions $\downarrow$}  & \textbf{F-score $\uparrow$}  \\
\hline
One-Level Attention & 37.33 & 60.05 \\
\hline
Two-Level Attention & 32.45 & 65.75 \\
\hline
Multimodal H-RNN \cite{sanabria2019deep} & 36.73 & 63.08 \\
\hline
HMA & 27.31 & 70.31 \\
\hline
\end{tabular}
\end{center}
\end{table}

Table \ref{tab:attention} shows the Missing Actions rate and F-score of the aforementioned methods. Our method misses at least 5\% less actions and gets an increase of 7\% in F-score. We believe our model outperforms \textit{Two-Level Attention} because this method does the multimodal fusion at the action level. Learning the importance of the event using only the audio features of a soccer match is a very difficult task.

\textbf{Comparison with frame-based models.} As mentioned in Section \ref{sec:related_work}, most of summarization methods are not suitable to compare with ours. Some methods \cite{zhang2016summary, zhang2016video, mahasseni2017unsupervised, zhang2018retrospective, gygli2015video, li2017general} are based on the optimization of summary diversity which is not convenient for soccer videos since a summary could contain several similar actions. Other approaches \cite{fiao2016automatic, tang2012epicplay, chakrabarti2011event, merler2017automatic, shukla2018automatic, tang2018autohighlight} use different input multimedia data (text or comments from social networks) which are not easily reachable.

vsLSTM \cite{zhang2016video} and H-RNN \cite{zhao2017hierarchical} are summarization methods close to what we propose since input samples are full matches and optimization does not rely on diversity. Even though these methods were originally proposed to use frame features as input we also make some modifications to train them with our event data features. 

H-RNN splits the video on fixed-size segments and use as input frame features extracted from GoogleNet. We set the segment size to 10 and a subsampling of 2 frames-per-second. For the event-based experiment we have empirically found that it is better to choose a significantly smaller size, most likely because the mean number of events on a ground truth clip is 7. For this reason and to perform a fair comparison, we have used as segment size the bag size used in the Generation of Action Proposals stage.

vsLSTM is a bidirectional-LSTM followed by a multi-layer perceptron. The inputs are frame features extracted from GoogleNet with a sub-sampling of 2 frames-per-second.

\begin{table}[!h]
\begin{center}
\caption{Performance comparison with frames based models.}
\label{tab:frames_comparison}
\setlength{\tabcolsep}{5pt}
\begin{tabular}{|c|c|c|c|c|c|c|}
\hline
\multirow{2}{*}{\textbf{\thead{ Method}}}
 & \multicolumn{2}{c|}{\textbf{Precision}}  & \multicolumn{2}{c|}{\textbf{Recall}} &
 \multicolumn{2}{c|}{\textbf{F-score}} \\
\cline{2-7}
& \thead{Events} & \thead{Frames} & \thead{Events} & \thead{Frames} & \thead{Events} & \thead{Frames} \\
\hline
vsLSTM & 72.08 & 24.24 & 63.44 & 32.88 & 67.49 & 27.91 \\
\hline
H-RNN  & 54.33 & 54.73 & 50.00 & 41.10 & 52.07 & 46.94 \\
\hline
\end{tabular}
\end{center}
\end{table}

For both methods trained with event features, the input is the same metadata feature vector $x_{e_f}^M$ used for our algorithm. 
In order to verify that the use of event data is more optimal than frames, Table \ref{tab:frames_comparison} provides the precision, recall and F-score for the methods trained with frame features and event data. Both vsLSTM and H-RNN perform better using events. In addition, the training and inference time when using frames is at least 10 times higher. 

The state-of-the-art results show that H-RNN usually performs better than vsLSTM \cite{zhang2018retrospective, zhao2017hierarchical, zhao2018hsa}, however with our data vsLSTM obtains better scores than H-RNN. Probably the fixed size of the segment and the overlap to decide if the segment is positive, have to be carefully analyzed. We have tried different segment sizes and overlap ratio and report the best results.

\begin{table}
\begin{center}
\caption{Performance comparison of Separate Modalities.}\label{tab:modalities}
\setlength{\tabcolsep}{5pt}
\begin{tabular}{|c|c|c|c|}
\hline
\textbf{Method} & \textbf{Missing Actions $\downarrow$} & \textbf{F-score $\uparrow$}\\
\hline
Only Audio& 24.91 & 61.83 \\
\hline
Only Metadata & 29.45 & 68.90 \\
\hline
HMA & 27.31 & 70.31 \\
\hline
\end{tabular}
\end{center}
\end{table}

\textbf{Multimodality. }We also evaluate the performance of each modality separately. We train an LSTM with an attention layer for the audio and another one for the metadata, using the $x_{e_f}^A$ and $x_{e_f}^M$ vector features respectively.
Table \ref{tab:modalities} shows that our method obtains the highest F-score compared with the models using only audio and only metadata features. Although using only audio features less actions are missing, the low F-score reveals the low precision of this method since it predicts a lot of false positives. This behavior is expected in sports videos since the crowd might produce a lot of noise even when an action is not important enough to be part of the summary. Comparing our results with the method using only metadata we can see that adding the audio features helps to reduce almost 2\% of missing actions.

\textbf{Soccer Baselines}.Most of the state-of-the-art methods on sports summarization evaluate their performance based on the detection of the most common actions such as goals or shots on target. We propose three different baselines to do a fair comparison and also to ensure that the summaries of our datasets do not follow a rule-specific pattern. 

\begin{itemize}[leftmargin=*]
     \item \textit{Random}: The prediction is a value taken from a continuous uniform distribution over the interval $[0,1)$, where the samples with values below 0.5 are negatives and the ones greater or equal than 0.5 are positives.
    \item \textit{Goals}: Since the easiest way to create a summary from a soccer video is to extract the goals of the match. This baseline considers as positive only the goal actions.
    \item \textit{Shots-on-Target}: As the goals are not enough to create a soccer summary, this baseline broadens the type of actions considered as positive. All Shots on Target actions (i.e.,goals, goalkeeper saving a shot on goal, any shot which goes wide or over the goal and whenever the ball hits the frame of the goal) are predicted as positive.
\end{itemize} 

\begin{table}[!h]
\begin{center}
\caption{Performance comparison of Soccer Baselines.}\label{tab:soccer_baselines}
\setlength{\tabcolsep}{5pt}
\begin{tabular}{|c|c|c|c|c|}
\hline
\textbf{Method} & \textbf{Precision} & \textbf{Recall} & \textbf{F-score}\\
\hline
Only Goals & 98.94 & 24.06 & 38.71 \\ 
\hline
All Shots-on-Target & 43.27	& 74.06 & 54.63 \\
\hline
Random & 41.37 & 43.49	& 42.40\\
\hline
HMA & 68.08 & 72.69 & 70.31 \\ 
\hline
\end{tabular}
\end{center}
\end{table}

Table \ref{tab:soccer_baselines} compares the performance of these baselines and our method. Our F-score is clearly the highest, outperforming at least 15\% the second best. The baseline \textit{Goals} gets a precision score near to the maximum because it is very common that all the goals of the match belong to the summary, however its recall is the lowest since it misses many other type of actions. The recall of our approach is only outperformed by \textit{All Shots-on-Target} since the type of actions considered in this baseline represent a big percentage of the actions generally included in summaries, yet its precision is at least 24\% lower than ours, hence this baseline predicts more false positives than us. It can be interpreted as our algorithm manages to extract some knowledge to predict in a cleverer way which shot actions should be in the summary.

\subsection{Multiple Summaries Generation}

In order to generate multiple summaries from the same match, we sample ten times from the $pl$ distribution shown in Equation \ref{eq:pl_sampling} with $\sigma = 0.05$, where $\theta$ are the output values of HMA stage. 

Each sample taken from $pl$ generates a ranked list of actions per video, where the ranking represents the importance of the action. Each of these lists is used to create a candidate summary in the following way: The action with the highest rank is selected to be part of the candidate summary and it is removed from the ranked list, then the procedure is repeated until the candidate summary has a duration that is very close to the ground-truth video summary duration.

\begin{table}[!h]
\begin{center}
\caption{Ranking Example. Test ranking results of a match in the Premier League dataset. The rows are ordered by time in the match. The \textcolor{red}{x} symbol indicates there is a ground truth action inside that time interval which was not predicted by the Generation of Action Proposals stage. The symbol $\bullet$ in column \textit{Proposals} shows if the was predicted as proposal. The numbers in the $pl$ columns indicate the position of the action in each generated ranking. Blank space in the $pl$ columns means that the action does not belong to the generated summary.}
\label{tab:ranking_example}
\setlength{\tabcolsep}{2.5pt}
\begin{tabular}{|c|c|c|c|c|c|c|c|c|c|c|c|c|}
\hline
Time&\textbf{gt} & \textbf{Proposals} & \textbf{$pl_1$}& \textbf{$pl_2$}& \textbf{$pl_3$}& \textbf{$pl_4$}& \textbf{$pl_5$}& \textbf{$pl_6$}& \textbf{$pl_7$}& \textbf{$pl_8$}& \textbf{$pl_9$}& \textbf{$pl_{10}$}\\
\hline
0'&start & $\bullet$ & 1 & 3 & 1 & 1 & 2 & 3 & 2 & 2 & 3 & 3  \\
\hline
19'&corner & $\bullet$ & 13 & 12 & & & 12 & 12 & 10 & 13 & 12 & 13\\
\hline
22'& & shot & 10 & 9 & 10 & 10 & 10 & 11 & 13 & 10 & 10 & 11\\
\hline
23'&shot & $\bullet$ & 4 & 4 & 5 & 4 & 4 & 6 & 4 & 4 & 4 & 4\\
\hline
26'& save & \textcolor{red}{x} & \textcolor{red}{x} & \textcolor{red}{x} & \textcolor{red}{x} & \textcolor{red}{x} & \textcolor{red}{x} & \textcolor{red}{x} & \textcolor{red}{x} & \textcolor{red}{x} & \textcolor{red}{x} & \textcolor{red}{x}\\
\hline
26'& & save & 6 & 5 & 4 & 5 & 5 & 5 & 6 & 5 & 5 & 5\\
\hline
29'& & free-kick &  &  & 11 &  &  &  &  &  &  & 10\\
\hline
30'&shot & $\bullet$ & 9 & 8 & 9 & 9 & 9 & 9 & 8 & 9 & 9 & 9\\
\hline
39'& & shot & 12 & 11 &  & 12 &  & 13 & 12 & 11 & 13 & \\
\hline
51'& save & $\bullet$ & 5 & 6 & 6 & 6 & 6 & 5 & 5 & 6 & 6 & 6\\
\hline
51'& & corner & 11 & 10 & 12 & 11 & 11 & 10 & 11 & 12 & 11 & 12\\
\hline
57'&goal & $\bullet$ & 2 & 2 & 3 & 2 & 3 & 1 & 1 & 3 & 1 & 2\\
\hline
66'& corner & \textcolor{red}{x} & \textcolor{red}{x} & \textcolor{red}{x} & \textcolor{red}{x} & \textcolor{red}{x} & \textcolor{red}{x} & \textcolor{red}{x} & \textcolor{red}{x} & \textcolor{red}{x} & \textcolor{red}{x} & \textcolor{red}{x}\\
\hline
67'& & save & 7 & 8 & 7 & 8 & 8 & 7 & 9 & 8 & 8 & 7 \\
\hline
83'&save & $\bullet$ &  & 7 & 8 & 7 & 7 & 8 & 7 & 7 & 7 & 8 \\
\hline
85'& & shot & 8 &  &  &  & 13 &  &  &  &  &  \\
\hline
93'&end & $\bullet$ & 3 & 1 & 2 & 3 & 1 & 2 & 3 & 1 & 2 & 1 \\
\hline
\end{tabular}
\end{center}
\end{table}

Let's explain more in detail the generation of multiple summaries from a specific example of our dataset. Table \ref{tab:ranking_example} shows the output of the 10 rankings generated by our approach for one of the Premier League matches. The Generation of Action Proposals stage predicted 19 action proposals for this match, which are described in the column \textit{Proposals}. The table lists 21 because the two ground-truth actions (lines 26' and 66' of the table) that were not correctly classified are also included for illustration purposes. The HMA network outputs a score for each of these 19 actions to then use it as $\theta$ for the Plackett-Luce distribution. 

The generated summary using $pl_6$ ranking contains 13 actions, which means that if we sum-up the duration of the 13 most important actions according to this ranking, we will get a video summary with a duration very close to the ground-truth one. This ranking considers the goal action as the most important, followed by the end and start of the match, since they are ranked 1,2,3 respectively. 

As a general analysis of this table, we can see that even though a save action (upper 26' of the table) was not detected as an action in the first stage of our algorithm, there was another save action (lower 26' of the table) in a near time which belongs to all the generated summaries. The summary generated from ranking $pl_1$ does not contain the last save action (line 85' of the table) of the ground-truth summary but it has another save action (line 83' of the table) from earlier in the match. 

\begin{table}[!h]
\begin{center}
\caption{Performance comparison of Multiple Summaries Generation.}
\label{tab:ranking_baselines}
\setlength{\tabcolsep}{5pt}
\begin{tabular}{|c|c|c|c|}
\hline
\textbf{Method} & \textbf{Missing Actions $\downarrow$} & \textbf{F-score $\uparrow$}\\
\hline
Random Ranking & 15.67 & 72.43 \\
\hline
Collyda et al \cite{collyda2020web} & 6.67 & 78.36 \\
\hline
Ours & 2.91 & 85.07 \\
\hline
\end{tabular}
\end{center}
\end{table}

We define two baselines to compare with our method. The first one is based on the method proposed by Collyda et al. \cite{collyda2020web}, which ranks the output of a summarization method from higher to lower score. We rank the score generated by HMA stage. The second one is a \textit{Random Ranking} where we shuffle the list of actions and the ranking is the position of the action in this shuffled list. 
The comparison results are depicted in Table \ref{tab:ranking_baselines}. We choose the best ranking among 10 using the validation set and report the results of that ranking in the test set. Our method misses less than 3\% of the actions and gets at least 6\% higher F-score than the state-of-the art method. 

An important advantage of our method is that it provides a possible solution to the always-present subjectivity in the summarization of sports videos. We provide multiple summaries to the final user that are close enough to not lose important information of the match but different enough to have options to choose. For instance, Table \ref{tab:ranking_example} shows that a free-kick action is in some of the generated summaries, adding some variability to the options without adding irrelevant actions.

\subsection{Transfer Learning}

\begin{table}
\begin{center}
\caption{Transfer Learning performance for the Summarization of the World Cup 2018 dataset.}
\label{tab:wc_ranking}
\setlength{\tabcolsep}{5pt}
\begin{tabular}{|c|c|c|c|c|}
\hline
\textbf{Method} & \textbf{Precision} & \textbf{Recall}& \textbf{F-score}\\
\hline
HMA & 38.58 & 72.05 & 50.25 \\
\hline
\thead{Multiple Summaries Generation} & 93.79 & 87.03 & 90.28 \\
\hline
\end{tabular}
\end{center}
\end{table}

As mentioned before, to the extent of our knowledge there is no available dataset for soccer summarization. However, Pappalardo et al. \cite{pappalardo2019playerank} released a set of soccer-logs collected by Wyscout, containing all the spatio-temporal events that occur during all matches of an entire season of seven competitions. We choose the World Cup 2018 data since it is the competition we could find online the most number of matches and summaries videos.

We follow the same steps described in Section \ref{sec:experiments} for the Premier League dataset. The code provided by Decroos et al. \cite{decroos2019actions} was very useful to homogenize the two different sources of event data. Then we predict using the models trained with the Premier League data without doing any kind of fine-tuning. Both LSTM-MIL and HMA models were trained only with the Premier League dataset.

It is important to emphasize that there are many differences between the two competitions. In terms of audio, the Premier League commentators are French while in the World Cup videos are English, which is unofficially known to have a different level of excitement in the match coverage. Also, the crowd cheering is not the same in a league than in a world cup competition. For the metadata, Premier League matches are better detailed, so they have more events.

Table \ref{tab:wc_ranking} shows the results obtained by the transfer learning of the HMA model. The Recall is high, but the Precision is very low which means that it outputs a lot of positive actions but many of them are not really part of the summary. This is an expected behavior since the video summaries of the World Cup are clearly shorter (around 2 minutes) than the ones of the Premier League (around 4 minutes). Usually when a summary is longer, it contains more actions like additional goal opportunities. Therefore, the model is trained to add more actions to the summary. 

Even though this duration difference negatively affects the summarization scores, it is the perfect condition for our generation of multiple summaries. Those additional actions with high prediction score in the HMA model become good candidates to generate other summaries. The precision is improved by 55\%, the Recall by 15\% and the F-score by 40\%. This confirms that our method can provide to the final user reliable summaries for the same match, even if the models were trained with a very different competition. 

\begin{table}
\begin{center}
\caption{Performance comparison of Multiple Summaries Generation with the World Cup 2018 dataset.}
\label{tab:wc_ranking_baselines}
\setlength{\tabcolsep}{5pt}
\begin{tabular}{|c|c|c|c|}
\hline
\textbf{Method} & \textbf{Missing Actions $\downarrow$} & \textbf{F-score $\uparrow$}\\
\hline
Random Ranking & 11.81 & 55.54 \\
\hline
Collyda et al \cite{collyda2020web} & 12.39 & 81.39 \\
\hline
Ours & 12.98 & 90.28 \\
\hline
\end{tabular}
\end{center}
\end{table}

In order to confirm our method also outperforms the baselines using the transfer learning on the World Cup dataset, Table \ref{tab:wc_ranking_baselines} compares the results with the previously described baselines. We outperform the Collyda et al. method by almost 9\%.

\section{Conclusion}
In this article, we have proposed a method for automatic generation of soccer videos based on event data and audio features. The proposed algorithm consists of three consecutive stages: a Generation of Action Proposals stage where we describe a new Multiple Instance Learning for sequentially dependent instances, a Multimodal Summarization stage that exploits event features and audio features through a novel hierarchical attention at event level instead of action level, and finally a generator of multiple summaries, based on Plackett-Luce model, tackles subjectivity and time-budget constraints of sports video summarization
. Experiments show that the three stages outperform state-of-the art methods and prove the generalizability of our model which can learn from one competition and transfer the knowledge to another competition acquired in different conditions and targeting different summary lengths. 

There are several key contribution factors: the idea of using event data for soccer video summarization, a multiple instance learning model for sequential data, the multimodal attention per event instead of per action and fitting a ranking distribution on the data to generate multiple summaries per match. 

Our method is not restricted to soccer since all the aforementioned companies providing precise event metadata are already providing similar information for several sports with large audiences (basketball, ice hockey, tennis, rugby, american football, among others). We will investigate in the future how easily our method can be transferred to other sports and at what cost.

\section*{Acknowledgment}
This work has been partially supported by Région Provence Alpes Côte d'Azur (PACA), by Université Côte d'Azur (UCA), by European Horizon 2020 research and innovation program (grant number 951911 - AI4Media), by French program, Investment in the Future, 3IA Cote d’Azur, (reference number ANR-19-P3IA-0002), and by Wildmoka Company.

\ifCLASSOPTIONcaptionsoff
  \newpage
\fi



%

\bibliographystyle{Transactions-Bibliography/IEEEtranS} 
\bibliography{ours}

\end{document}